\documentclass{article}
\usepackage{spconf,amsmath,graphicx}
\usepackage{float}
\usepackage{enumitem}
\setlist{nosep, leftmargin=14pt}
\usepackage[justification=centering]{caption}
\usepackage{mwe} 

\title{Human Not in the Loop: Objective Sample Difficulty Measures for Curriculum Learning}

\name{Zhengbo Zhou$^{\star}$ \qquad  Jun Luo$^{\star}$ \qquad  Dooman Arefan$^{\dagger}$ \qquad Gene Kitamura$^{\ddagger}$ \qquad Shandong Wu$^{\star \dagger \mathsection}$}

\address{$^{\star}$Intelligent Systems Program, University of Pittsburgh, Pittsburgh, PA, USA\\
    $^{\dagger}$Department of Radiology, University of Pittsburgh, Pittsburgh, PA, USA\\
    $^{\ddagger}$Department of Radiology, Loma Linda University, Loma Linda, CA, USA\\
    $^{\mathsection}$Department of Biomedical Informatics and Department of Bioengineering\\University of Pittsburgh, Pittsburgh, PA, USA}

\begin{document}

%
\maketitle
\begin{abstract}
Curriculum learning is a learning method that trains models in a meaningful order from easier to harder samples. A key here is to devise automatic and objective difficulty measures of samples. In the medical domain, previous work applied domain knowledge from human experts to qualitatively assess classification difficulty of medical images to guide curriculum learning, which requires extra annotation efforts, relies on subjective human experience, and may introduce bias. In this work, we propose a new automated curriculum learning technique using the variance of gradients (VoG) to compute an objective difficulty measure of samples and evaluated its effects on elbow fracture classification from X-ray images. Specifically, we used VoG as a metric to rank each sample in terms of the classification difficulty, where high VoG scores indicate more difficult cases for classification, to guide the curriculum training process We compared the proposed technique to a baseline (without curriculum learning), a previous method that used human annotations on classification difficulty, and anti-curriculum learning. Our experiment results showed comparable and higher performance for the binary and multi-class bone fracture classification tasks.

\end{abstract}
\begin{keywords}
Curriculum learning, Variance of Gradient, Elbow fracture, Medical imaging, Classification
\end{keywords}
\section{Introduction}
\label{sec:intro}

Curriculum learning (CL) has been proposed as a training strategy in deep learning to expedite learning and improve the model performance. In this technique, samples are presented in the order of from “easy” to “hard” to the model, instead of randomly selecting samples [1]. This is motivated by the learning process of humans and animals, who start learning with an easy ‘concept’ before gradually increasing the difficulties of the task. It has been shown that starting training with “easy” samples enables machine learning models to converge faster and shows a significant improvement in the generalization of the models \cite{bengio2009curriculum}. Due to those benefits, CL has been implemented in various machine learning tasks such as image classification  \cite{lotter2017multi}, object detection \cite{zhang2019leveraging}, multi-task learning \cite{sarafianos2018curriculum}, etc. The existing CL methods can be divided into predefined CL or automated CL,  in terms of assigning the difficulty  by humans or algorithms  \cite{wang2021survey}.

Previous studies in CL for medical image analysis are mostly based on predefined CL. Amelia et al. proposed strategies to assign a degree of difficulty based on human knowledge such as medical decision trees and inconsistencies between the annotations of multiple clinical experts \cite{jimenez2019medical}. Jun et al. proposed a strategy based on clinically known knowledge from human experts who qualitatively assigned the difficulty level of diagnosis for each elbow fracture subtype \cite{luo2021medical}.	Although both strategies improved the model performance by using the medical knowledge guided curriculum learning, they need extra efforts from human experts to annotate the difficulty level, which is subjective, qualitative, less reproducible, and may introduce bias.

\begin{table*}[ht]
\vspace{-4em}
\centering
\caption{Distribution of the normal and different elbow fracture subtypes in our dataset}
\begin{tabular}{llllllllll}
\hline
Type  & normal & \begin{tabular}[c]{@{}l@{}}Ulnar\\ Fracture\end{tabular} & \begin{tabular}[c]{@{}l@{}}Radial\\ Fracture\end{tabular} & \begin{tabular}[c]{@{}l@{}}Humeral\\ Fracture\end{tabular} & \begin{tabular}[c]{@{}l@{}}Dislocation\\ Fracture\end{tabular} & \begin{tabular}[c]{@{}l@{}}Complex \\ Fracture\end{tabular} & \begin{tabular}[c]{@{}l@{}}Coronoid \\ Fracture\end{tabular} & \begin{tabular}[c]{@{}l@{}}Total of \\ fracture\end{tabular} & Total \\ \hline
Train & 800    & 88                                                       & 340                                                       & 84                                                         & 11                                                             & 42                                                          & 27                                                           & 592                                                          & 1,392 \\
Test  & 400    & 10                                                       & 44                                                        & 9                                                          & 2                                                              & 4                                                           & 4                                                            & 73                                                           & 473   \\
Total & 1,200  & 98                                                       & 384                                                       & 93                                                         & 13                                                             & 46                                                          & 31                                                           & 665                                                          & 1,865 \\ \hline
\end{tabular}
\vspace{-1.5em}
\end{table*}
			
In this work, we developed a novel and automated CL strategy using variance of gradients (VoG) as a scoring criterion for classification difficulty. Here, samples with higher VoG scores indicate more difficult to classify, since images with a high VoG score are more likely to have clutter and complex patterns, making it more difficult for a deep learning model to classify the images [8]. We implemented a learning process using VoG ranking through permutations of the training epochs by sampling without replacement at the beginning, and then an algorithm was used to update the probability, until it reached a random shuffle status. We performed experiments on an elbow X-ray dataset of 1865 patients, by conducting binary classification (normal vs. fracture) and multi-class classification (normal vs. $3$ individual fracture subtypes). We compared the proposed technique to a baseline (without curriculum learning), a previous method that used human annotations on classification difficulty, and anti-curriculum learning. Results show that our method outperforms the compared methods and because of its automatic nature, it is objective, reproducible, and independent of human input. Our contributions can be summarized as follows:
	
\begin{itemize}
\item We adapted the VoG method from computer vision of natural image classification to the curriculum learning context and showed it is effective for medical imaging applications.
\item We showed in elbow fracture classification tasks that the VoG-based curriculum learning can either achieve comparable or higher performance than several compared methods, and our method is more efficient because it is an automated process and independent of human input.
\item We showed that the VoG-based objective scores and the human radiologists’ subjective scores measure different aspects of the difficulty of the classification samples, while they both can guide curriculum learning.
\end{itemize}
The rest of the paper is organized as follows. Section 2 describes the methods of our study. Section 3 presents experiment designs and datasets. Results are shown in Section 4. Section 5 discusses conclusions and future work.

\section{Methods}
\label{sec:format}
To design a curriculum paradigm in deep curriculum learning, two major components need to be determined. First, how to measure the training data and assign or derive a metric of sample difficulty. Second, how to present the measured samples in the training procedure \cite{wang2021survey}. In the following, we describe the difficulty measurer and training scheduler to address the two components.

\subsection{Difficulty Measurer}
\label{ssec:subhead}
The input image $X$ can be divided into individual pixels, referred to as $x_i$, with $i$ representing the index ranging from $1$ to $N$, where $N$ is the total number of pixels in the image. For the given image, the gradient of the activation $A$ for each pixel is computed where $l$ denotes as the pre-softmax layer of the network and $p$ is the index of either the true or predicted class probability. Let $S$ be a matrix that represents the gradient of $A^{l}_{p}$ concerning individual pixels $x_i$, and $S$ is calculated by Equation \eqref{1}. The gradient matrix $S$ will have the same size as an image \cite{agarwal2022estimating}.  
\begin{equation}
  S=\frac{\partial A^{l}_{p}}{\partial x_i}.   
  \label{1}
\end{equation}
In order to better learn features during training \cite{shrikumar2017learning}, we take the average over the color channels and the generated matrix $T$ records the gradient of each pixel $x_i$. In the next step, it shows how the variance of gradients are computed. We generate the above gradient matrix $T$ for all individual checkpoints in a given set of $K$ checkpoints to get gradients at different steps, i.e., ${T_1,...,T_K}$. We then calculate the mean gradient $\mu$ by by taking the average of the $K$ gradient matrices. Note, $\mu$ is the mean across different checkpoints and is of the same size as the gradient matrix $T$. Then the variance of gradients across each pixel is calculated as following \cite{agarwal2022estimating}:
\begin{equation}
  \mu = \frac{1}{K}\sum_{i=1}^{K}T_i.   
  \label{2}
\end{equation}
\begin{equation}
  VoG_p = \sqrt{\frac{1}{K}}\sum_{i=1}^{K}(T_i-\mu)^{2}.   
  \label{3}
\end{equation}
Once we have variance of gradients for each pixel, we then average the pixel-wise variance of gradients to compute the VoG score for the given input images and obtain the ranking of training samples based on VoG score.
\begin{table*}[htb]
\vspace{-4em}
\centering
\caption{Model performance on binary (normal vs. fracture) classification.\\ RS-CL: using radiologist-assigned difficulty score. VoG-CL: using VoG score. }

\begin{tabular}{llllllllll}
\hline
\multicolumn{2}{l}{}            & \multicolumn{2}{c}{Accuracy} & \multicolumn{2}{c}{Recall} & \multicolumn{2}{c}{AUC}  & \multicolumn{2}{c}{F1 Score} \\ \cline{3-10} 
\multicolumn{2}{l}{}            & Mean              & Std      & Mean             & Std     & Mean            & Std    & Mean              & Std      \\ \hline
\multicolumn{2}{l}{Baseline}    & 0.7497            & 0.0164   & 0.7517           & 0.0084  & 0.8407          & 0.0095 & 0.7208            & 0.0052   \\
\multicolumn{2}{l}{RS-CL}       & 0.7850            & 0.0106   & 0.7877           & 0.0076  & 0.8642          & 0.0109 & 0.7595            & 0.0103   \\
\multicolumn{2}{l}{VoG-CL}      & \textbf{0.7858}   & 0.0041   & \textbf{0.7885}  & 0.0060  & \textbf{0.8652} & 0.0099 & \textbf{0.7605}   & 0.0087   \\
\multicolumn{2}{l}{Anti-VoG-CL} & 0.7610            & 0.0201   & 0.7570           & 0.0116  & 0.8356          & 0.0148 & 0.7206            & 0.0148   \\ \hline
\end{tabular}
\end{table*}

\begin{table*}[htb]

\centering
\caption{Model performance on multi-class classification.}

\begin{tabular}{llllllll}

\hline
\multicolumn{2}{l}{}            & \multicolumn{2}{c}{Accuracy}                       & \multicolumn{2}{c}{Balanced Accuracy}              & \multicolumn{2}{c}{AUC}                            \\ \cline{3-8} 
\multicolumn{2}{l}{}            & \multicolumn{1}{c}{Mean} & \multicolumn{1}{c}{Std} & \multicolumn{1}{c}{Mean} & \multicolumn{1}{c}{Std} & Mean                     & Std                     \\ \hline
\multicolumn{2}{l}{Baseline}    & 0.7175                   & 0.0277                  & 0.5841                   & 0.0213                  & 0.8369                   & 0.0097                  \\
\multicolumn{2}{l}{RS-CL}       & 0.7322                   & 0.0200                  & 0.5613                   & 0.0364                  & 0.8496                   & 0.0098                  \\
\multicolumn{2}{l}{VoG-CL}      & \textbf{0.7337}          & 0.0170                  & \textbf{0.6294}          & 0.0625                  & \textbf{0.8570}          & 0.0107                  \\
\multicolumn{2}{l}{Anti-VoG-CL} & 0.7227                   & 0.0279                  & 0.5775                   & 0.0374                  & 0.8417                   & 0.0095                  \\ \hline
\multicolumn{2}{l}{}            & \multicolumn{2}{c}{Binary Accuracy}                & \multicolumn{2}{c}{Binary AUC}                     & \multicolumn{2}{c}{F1 Score}                       \\ \cline{3-8} 
\multicolumn{2}{l}{}            & \multicolumn{1}{c}{Mean} & \multicolumn{1}{c}{Std} & \multicolumn{1}{c}{Mean} & \multicolumn{1}{c}{Std} & \multicolumn{1}{c}{Mean} & \multicolumn{1}{c}{Std} \\ \hline
\multicolumn{2}{l}{Baseline}    & 0.7678                   & 0.0368                  & 0.8325                   & 0.0154                  & 0.7139                   & 0.0269                  \\
\multicolumn{2}{l}{RS-CL}       & 0.7715                   & 0.0152                  & 0.8386                   & 0.0156                  & 0.7246                   & 0.0167                  \\
\multicolumn{2}{l}{VoG-CL}      & \textbf{0.7853}          & 0.0265                  & \textbf{0.8527}          & 0.0104                  & \textbf{0.7358}                   & 0.0203                  \\
\multicolumn{2}{l}{Anti-VoG-CL} & 0.7620                   & 0.0199                  & 0.8377                   & 0.0132                  & 0.7174                   & 0.0211                  \\ \hline
\end{tabular}
\vspace{-1.5em}
\end{table*}

\subsection{Training Schedular}
\label{ssec:subhead}
Once the difficulty score is computed, a training schedular is then applied during the training based on the VoG difficulty score. Our curriculum learning method is equipped with a sampling-without-replacement strategy to permute and reorder the training data at the beginning of each epoch. Each training sample $s_i$ has a probability, $p_i$, and its value is updated at the beginning of each training epoch in order to implement curriculum learning. Each sample is first assigned with the initial value $p_{i,(1)}$ at the first epoch at initial and then updated based on the updating algorithm. Finally, after certain epochs, the data are randomly shuffled and each sample shares the same probability. 

For the initialization of $p_i$, the $p_i$ of all images is calculated based on their VoG ranking scores. These scores reflect the level of difficulty in learning the features of the samples during the training process. To distinguish the importance of ‘easy’ and ‘hard’ samples, easier samples with lower VoG scores are assigned with a higher probability according to according to Equation \eqref{4}:
\begin{equation}
  p_{i,(1)} =  \frac{s_{i}}{\sum_{j=1}^{N}s_{j}},  
\label{4}
\end{equation}

\noindent where $s_k$ represents the rank of the sample $x_k$ in the sorted VoG scores of all training samples from highest to lowest, and $N$ denotes the number of images in the training set. 
	 	 	 		
After the first epoch, the sampling probability of all the training samples is updated and we denote the updated sample probability for each $x_i$ at epoch $e$ as $p_{i,(e)}$, and $\lambda_i$ (Equation \eqref{5}) is a scalar for updating sampling probability of $xi$. $p_{i,(e)}$ can be calculated by the sampling probability of the previous epoch $p_{i,(e-1)}$ multiple $\lambda_i$. This ensures $p_i$ for every image to exponentially and smoothly transit to $p_{i,(final)} = 1/N$. $L$ is the number of epochs after which $p_i$ equals to $p_{i,(final)}$; for the rest of the training process and after $L$ epoch, the sampling probability of each sample will not be updated. $L$ is treated as a hyperparameter for determining when the training reaches a stage of randomly shuffling the dataset. This way, the model can gradually reduce the emphases we initially given on the easier cases and then reach a stage of random selection of samples in the entire training set after epoch $L$ \cite{luo2021medical}. 
\begin{equation}
\label{5}
  \lambda_i = \sqrt[L]{\frac{1/N}{p_{i,(1)}}}.  
\end{equation}

\section{Experiment}
\label{sec:format}

\begin{figure}[htb]
\vspace{-4.0em}
\setlength{\abovecaptionskip}{0pt}
\setlength{\belowcaptionskip}{-0.5cm}
\begin{minipage}[b]{.96\linewidth}
  \centering
  \centerline{\includegraphics[width=8.0cm]{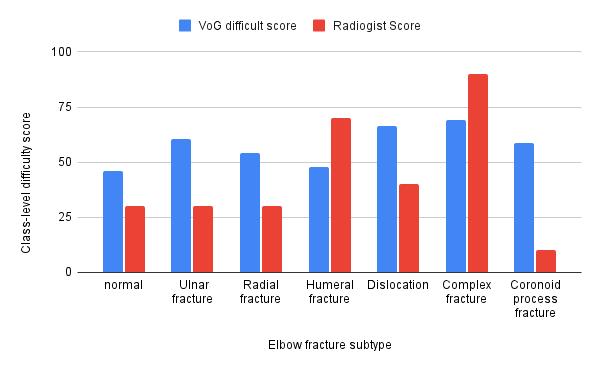}}
  \centerline{(a) Binary classification }\medskip
\end{minipage}
\hfill
\begin{minipage}[b]{0.96\linewidth}
  \centering
  \centerline{\includegraphics[width=8.0cm]{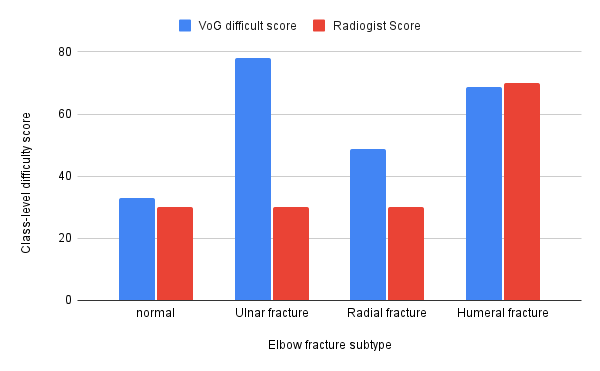}}
  \centerline{(b) Multi-classification }\medskip
\end{minipage}
\caption{Comparison between VoG-based and radiologist-assigned difficult scores at the class level for each fracture subtype.}
\label{fig:res}
\vspace{-0.5em}
\end{figure}
\textbf{Dataset.} The dataset used for evaluation of our proposed method contained a total of 1865 different X-ray elbow images retrospectively collected at our institution. There are 665 fracture-positive and 1200 non-fracture (normal) cases. All fracture-positive images were categorized by an experienced radiologist into six different fracture subtypes, that is, (a) Ulnar fracture, (b) Radial fracture, (c) Humeral fracture, (d) Dislocation, (e) Complex fracture and (f) Coronoid process fracture.

We defined two different experimental tasks in this study, including binary and multi-class classification. In the binary classification task, we merged all six fracture subtypes in one class as a fracture-positive cohort versus non-fracture (normal) cohort. We used a stratified technique based on different fracture subtypes to split the entire dataset into training and test sets. Table 1 displays the data distribution in the training and test set for each elbow fracture subtype. During the test process, we randomly split the 400 normal cases into 4 subsets, each containing one hundred images. We then evaluated our methods on an unseen subset of 100 normal cases and all 73 fracture cases to form a balanced test. We repeated the test 4 times, each time selecting a different set of normal cases, and the average performance results were reported.

In the multi-class classification task, images with Dislocation, Complex, and Coronoid fracture were excluded due to their very small sample sizes. Similar to the binary classification task, the test was repeated 4 times, and average results were reported.\\

\noindent \textbf{Metrics.} In the binary classification, accuracy, recall, Area Under the ROC Curve (AUC), and F1 score were used to evaluate models’ performance. In the multi-task classification task, we calculated accuracy, balanced accuracy, AUC, binary task accuracy, binary task AUC and F1 score. Balanced accuracy was calculated by averaging the ratios between the number of true positives and that of total samples in each class, which reduces the effect caused by class imbalance. Binary task accuracy and binary task AUC were computed by transforming the prediction to 'normal' and 'fracture'. If the ground truth and the predicted labels are 'normal', they are set to 0, and set to 1 otherwise. \\

\noindent \textbf{Implementation Details.} We employed the VGG16 convolutional neural network architecture \cite{simonyan2014very} to implement the classification models. We compared our proposed training strategy to three methods on both the binary and  multi-class classification tasks:
 1) Baseline model: it is based on randomly shuffling the training data for each epoch. 2) Radiologists Score (RS)-CL: This is a clinical knowledge-guided curriculum learning, which uses the difficulty score assigned by an experienced radiologist to each fracture subtype. We used the scores given in a prevous work \cite{luo2021medical}. 3) Anti-curriculum learning: This method reverses the ranking scores computed by VoG, and starts the training from the “difficult” samples. The model architecture and curriculums were implemented by PyTorch and ran on an NVIDIA TESLA V100 GPU from a local supercomputing center. All the hyperparameter tuning and data augmentation were the same for all experiments. The experiment results are summarized using the mean and standard deviation across the 5 runs. All models were run for 30 epochs using the cross-entropy loss. VoG was computed using checkpoints from the 26th, 28th, and 30th epochs, which were experimentally determined.

In addition, in order to gain insights on the VoG scores, we also compared the VoG difficult scores derived automatically from the samples to the difficulty scores given by the radiologist at the class level. Based on the VoG scores during training, the ranking of each training sample can be calculated at the instance level. To calculate the class-level difficulty score based on the instance level VoG scores, each training sample is assigned a difficult score $D_i$, as follows:
\begin{equation}
 D_i = \frac{(N - r_i)}{N} * 100, 
\label{7}
\end{equation}

\noindent where $r_i$ is the ranking of each sample and $N$ represents the number of the training samples. Then we computed the average of the difficulty scores across the training sample in each class to generate a class-level VoG difficult score, for each of the fracture subtype. We plot histograms to compare the distributions of VoG scores and radiologist-assigned scores to see how they may be related at the class level.

\section{Results}
\label{sec:pagestyle}

The main results are shown in Tables 2 and 3. As can be seen, in the binary classification task, VoG-based CL achieved comparable performance to the RS-CL method (Table 2), with a mean accuracy of 0.786 vs. 0.785 (RS-CL) and an AUC of 0.865 vs. 0.864 (RS-CL). Both the VoG-CL and RS-CL outperformed the baseline method and anti-curriculum method, as expected. Note that even though the performance is very close between VoG-CL and RS-CL, the proposed VoG method is superior as it is an automated curriculum and independent of human input. As shown in Table 3, our proposed VoG-based CL outperformed all the compared
methods (including RS-CL) in the multi-class classification task, showing a mean balanced accuracy of 0.629 and AUC of 0.857. This experiment shows advantages of VoG-CL over RS-CL in a more complex classification task. In addition, the anti-curriculum settings showed inferior results compared to the curriculum settings, as expected in both classification tasks.

Fig. 1 shows the comparisons between VoG-based class difficulty score and radiologist-assigned scores. In the binary classification task, class “normal” had the lowest difficulty score while class Complex fracture had the highest score, which was in line with the radiologist’s difficulty scores. For the multi-class classification task, no obvious patterns were found. One consideration here is the smaller sample sizes for class Ulnar fracture and Humeral fracture, which may have an influence to the VoG scores. Overall, from what we see in Fig. 1, there are no obvious matching between the two different types of scores, indicating they measure the samples in different aspects, while they both can guide curriculum learning, and achieved similar performance in the binary classification task. Further experiments are needed to evaluate this finding.

\section{Conclusion}
\label{sec:typestyle}
In this work, we proposed the VoG-based method to automatically measure sample difficulty for curriculum learning. In the elbow fracture subtype classification tasks, our method showed comparable or outperforming effects. Our method can save the efforts from human annotation on the difficulty score, prevent the subjectivity from human input and thus potential bias. While we focused on one specific application of fracture classification, our method can be easily extended to other diseases or medical imaging classification problems. One limitation of our study is the small sample sizes for several subtype fractures and the data are not well balanced. In future work, we will perform more evaluation using larger and other datasets as well as different imaging classification tasks.

\section{Compliance with Ethical Standards}
\label{sec:typestyle}

This retrospective single-center study was approved by Institutional Review Board (IRB) at our institution and was conducted compliant to Health Insurance Portability and Accountability Act (HIPAA). All medical data were de-identified before used for machine learning modeling.\\

\noindent \textbf{Acknowledgements}. This project was supported in part by a National Institutes of Health (NIH) grant (1R01CA218405), the grant 1R01EB032896 as part of the National Science Foundation (NSF)/NIH Smart Health and Biomedical Research in the Era of Artificial Intelligence and Advanced Data Science Program, a NSF grant (CICI: SIVD: \#2115082), and the University of Pittsburgh Momentum Funds (a scaling grant) for the Pittsburgh Center for AI Innovation in Medical Imaging. This work used the computational support by the Extreme Science and Engineering Discovery Environment (XSEDE), which is supported by NSF grant number ACI-1548562. Specifically, it uses the Bridges system, which is supported by NSF award number ACI-1445606, at the Pittsburgh Supercomputing Center.

\bibliographystyle{IEEEbib}
\bibliography{strings,refs}

\end{document}